\documentclass{amia}
\usepackage{graphicx}
\usepackage[labelfont=bf]{caption}
\usepackage[superscript,nomove]{cite}
\usepackage{color}
\usepackage[table,xcdraw]{xcolor}
\usepackage[colorinlistoftodos]{todonotes}
\usepackage{hyperref}
\usepackage{url}
\usepackage[utf8]{inputenc}
\usepackage{siunitx}
\usepackage[symbol]{footmisc}

\begin{document}

\title{On the explainability of hospitalization prediction on a large COVID-19 patient dataset}

\author{

Ivan Girardi, PhD$^1${\footnote[2]{Equal contributions}}, Panagiotis Vagenas, MS$^1$$^{\dagger}$,
Dario Arcos-Díaz, PhD$^2$, \\ Lydia Bessaï, MS$^2$, Alexander Büsser, MS$^3$,
Ludovico Furlan, MD$^4$, Raffaello Furlan, MD$^5$, Mauro Gatti, PhD$^6$,
Andrea Giovannini, PhD$^1$,
Ellen Hoeven, MS$^2$, Chiara Marchiori, PhD$^1$
}

\institutes{
     $^1$IBM Research Europe,
     $^2$IBM GBS Germany,
     $^3$IBM GBS Switzerland\\
     $^4$Fondazione IRCCS Ca’ Granda, Ospedale Maggiore Policlinico, Milano, Italy
     $^5$Department of Biomedical Sciences, Humanitas University and IRCCS - Humanitas Research Hospital, Milano, Italy,
     $^6$IBM GBS Italy
     }
 
\maketitle
\noindent{\bf Abstract}

\textit{We develop various AI models to predict hospitalization on a large (over 110$k$) cohort of COVID-19 positive-tested US patients, sourced from March 2020 to February 2021. Models range from Random Forest to Neural Network (NN) and Time Convolutional NN, where combination of the data modalities (tabular and time dependent) are performed at different stages ({\it early} vs. {\it model fusion}). Despite high data unbalance, the models reach average precision 0.96-0.98 (0.75-0.85), recall 0.96-0.98 (0.74-0.85), and $F_1$-score 0.97-0.98 (0.79-0.83) on the non-hospitalized (or hospitalized) class. Performances do not significantly drop even when selected lists of features are removed to study model adaptability to different scenarios. However, a systematic study of the SHAP feature importance values for the developed models in the different scenarios shows a large variability across models and use cases. This calls for even more complete studies on several explainability methods before their adoption in high-stakes scenarios.}

\section*{Introduction}
As recently published \cite{LancetV3J21}, during the COVID-19 pandemic, several AI-based models have been developed, targeting a large variety of application areas, such as identification of disease clusters, monitoring of cases, prediction of the future outbreaks, mortality risk, diagnosis of COVID-19, disease management, or vaccine development \cite{AroraFutVir2020}. A thorough review found that almost all prediction models were trained on small or low-quality datasets, poorly reported and at a very high risk of bias or overfitting \cite{WynantsBMJ2020}. Due to the sense of urgency and the lax regulatory landscape, some of these models were even used without clinical validation, raising legitimate concerns about patient safety and model clinical validity \cite{MythGeneralisability}. 

A way to build more robust AI prediction models is to adhere to the PROBAST framework, which provides a structured approach to assess potential risk of bias and applicability to the intended population and setting\cite{PROBAST}. Transparent reporting can be achieved by following the TRIPOD checklist\cite{TRIPOD}. While the validation of the model "as-is" on external datasets is perceived as a proof of the generalization power of the model by some, this may be very difficult due to intrinsic differences in available populations, collected data, protocols and guidelines. Instead, a definitive important step forward would be to validate the models in real clinical settings by SMEs while in parallel improving their interpretability and correcting their bias with the help of explainability methods. 

In this paper, we develop hospitalization prediction models on a large cohort of COVID-19 case data collected in the US from March 2020 to February 2021 and comprising more than 110$k$ positive-tested patients (113941 upon data extraction, 110996 after further preprocessing). Knowing which patients are at risk for hospitalization for COVID-19-related symptoms or complications can help physicians decide not only how to best manage a patient's care from the time of testing (e.g. remote monitoring, more frequent encounters), but also how to allocate resources. All the models reach relatively high performance despite the high class unbalance and the noise, typical of large real world datasets. We find however that despite the similar performance, different models lead to very different explanation results. While some hospitalization prediction models have been published \cite{Jimenez-SolemScienRep2021, ElifeScience, RichardsonJAMA2020,JehiPlos}, they are generally built on smaller cohorts and do not systematically explore explainability methods across different AI models and feature subsets.

\section*{Methods}
In this section we describe in details the various components of our approach: (i) Cohort definition; (ii) Data extraction: sourcing and preparation of data with partition of patient history in time intervals to enable increased flexibility during feature extraction;
(iii) Feature extraction: accurate preparation and selection of features with special attention to data leakage prevention, missing data imputation, and proper feature encoding for capturing the temporal dimension of the data; (iv) Model development, where we experiment with time dependent and independent models (early and model fusion of temporal and Boolean modalities), (v) Internal testing and adaptability of the model for the use in several settings, by reducing the number of features used to train the model or varying the time interval between predictor assessment and outcome determination, and (vi) Model explainability.

\subsection*{Cohort definition}
\label{section:cohorts}
Our analysis was performed on the IBM Explorys Electronic Health Record (EHR) Database \cite{Explorys}, which contains real-world clinical, operational, and financial data, spanning various healthcare aspects, from ambulatory to inpatient to specialty care. Explorys data are sourced from integrated delivery networks (IDN), clinically integrated network providers (CIN), and care collaborative groups (CC) and are constantly updated with incoming records. 
Data are curated, standardized, and normalized by using medical international coding standards, classification systems, ontologies, and lab unit measures such as ICD-10, SNOMED, LOINC, and RxNorm. Furthermore, data are searchable through a de-identified database. 
Explorys data have been used by the medical community for several studies, such as recently for building a predictive model for chronic kidney disease \cite{Ravizza}.

We used data collected from March 2020 to February 2021. To draw evidence of COVID-19 infection we used \textit{diagnoses} and \textit{observations}. We defined two evidence classes for COVID-19 diagnoses, \textit{confirmed} and \textit{suspected}, and examined the diagnosis fields \textit{ICD-10 code}, the \textit{SNOMED concept}, and a short \textit{diagnosis description} snippet available in our source EHR data. As \textit{confirmed} we defined the diagnoses explicitly registered with the ICD-10 code ``U07.1", which was introduced by the WHO in April 2020 for use ``when COVID-19 has been confirmed by laboratory testing irrespective of severity of clinical signs or symptoms". Diagnoses with the ICD-10 code ``U07.2" (for use "when COVID-19 is diagnosed clinically or epidemiological but laboratory testing is inconclusive or not available") were classified as {\it suspected}. Where the above COVID-19-specific codes were absent, we searched for a set of specific COVID-19-associated text patterns in the diagnosis description as well as in SNOMED concept name (SNOMED search was performed in a hierarchy-aware manner, as described later. Pattern matches were classified as suspected, unless detected by a set of exclusion patterns we defined, in which case the diagnoses were altogether disregarded from further analysis. To extract observation-based COVID-19 evidence, we used LOINC ``94500-6", which is the recommended code when (i) the gene or region being tested is not specified and a qualitative result is being reported (e.g., Detected/Not detected); or (ii) a single qualitative overall result based on a combination of individual test results is being reported, and also by far the most frequently occurring COVID-19 diagnostic test LOINC code in the data. Observations registered with LOINC code ``94500-6" and positive result were defined as \textit{positive-test} \footnote{Testing availability was very limited until March-April 2020.}.

\begin{figure}[h!]
  \includegraphics[scale=0.28]{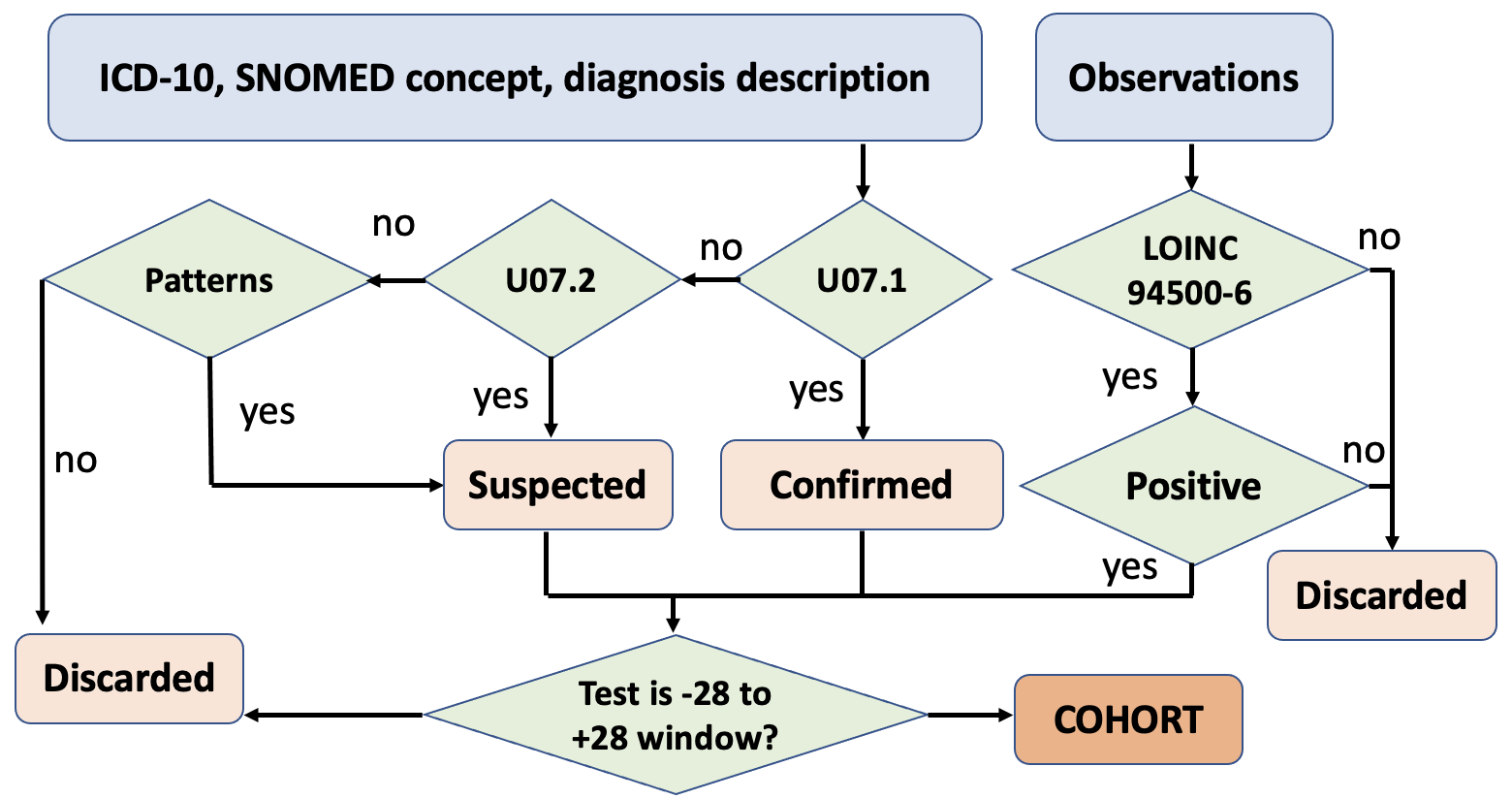}
 \vspace{0.2cm}
   \includegraphics[scale=0.17]{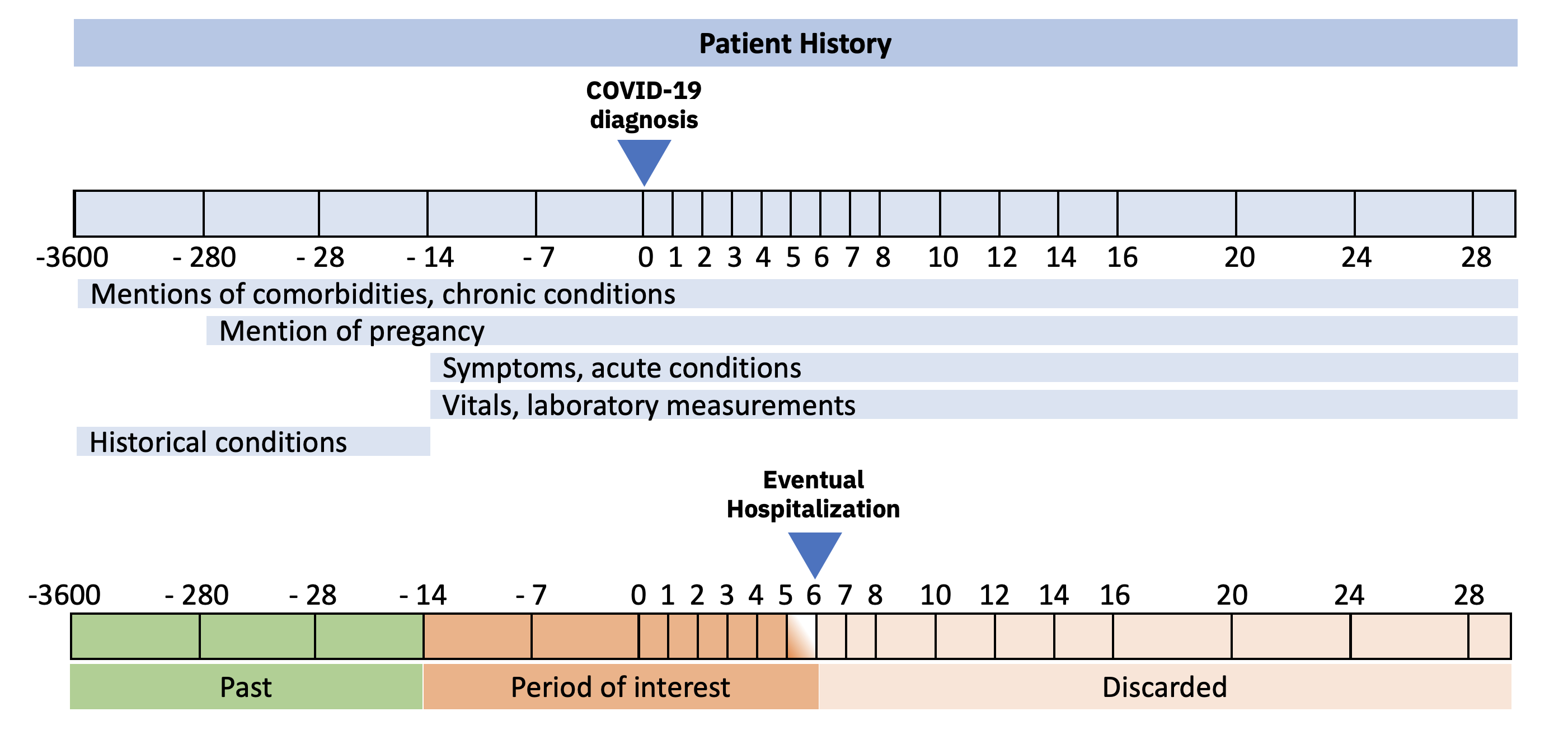}
 \caption{Left: cohort selection process. Right: Time partitioning of patient's history referenced to the COVID-19 diagnosis and collected features over
 time intervals (Blue). Selection of features over the period of interest for an exemplary patient hospitalized 6 days after the diagnosis (Colored regions).}
 \label{fig:Time}
  \vspace{0.5cm}
\end{figure}

As cohort of interest we defined the one consisting of patients who had both a COVID-19 diagnosis, whether {\it suspected} or {\it confirmed}, and a positive-test observation within a time frame from 4 weeks prior to their earliest COVID-19 diagnosis up to 4 weeks following it. This cohort finally comprised 113941 patients. A statistic summary can be seen on Table~\ref{table:stasboolean2}.
Restricting the cohort only to patients with {\it confirmed} COVID-19 diagnosis and positive-test observation in the same time frame as above did not lead to significantly different results. We excluded patients who had a COVID-19 diagnosis but no positive-test observation, in order to reduce noise and focus on cases documented more completely in our EHR data. The absence of an explicit positive-test, is indeed ambiguous and can be explained either by the patient having been diagnosed by other means, e.g. chest CT, or as missing information, e.g. case not properly documented, test done in other institutions or unavailable (selection bias was not studied in detail, but the demographic distribution when including such ``diagnosis-only" patients was similar to that of our defined cohort). A schema of the cohort
selection is given in Figure~\ref{fig:Time}.

\subsection*{Data extraction} 
For each patient belonging to the cohort, we extracted demographics (age, gender), medical conditions (chief complaints, symptoms, comorbidities), vital signs and laboratory measurements, based on literature findings and SME input. With the exception of the demographics, all pieces of information were extracted with their corresponding timestamp. To capture the temporal dimension of the data, but also to reduce complexity, each patient journey was partitioned into various time intervals using the patient's earliest COVID-19 diagnosis as reference point. 
Each variable was then extracted in an aggregated manner for each interval. The partitioning is depicted in Figure \ref{fig:Time} and was designed to allow a greater focus from the diagnosis time onward.

To maximize recall, the same approach based on SNOMED and text pattern matching used to extract the COVID-19 diagnosis was used to extract symptoms, comorbidities and other medical conditions. 
In the case of SNOMED search, we performed a hierarchical pattern matching including also matched ancestors and using exclusion patterns to reduce false positives.
We assigned a Boolean values to each medical condition in each time interval. Measurements and test results were extracted from the \textit{observation} data, using corresponding LOINC codes and were normalized into a single unit. For each observation we extracted one, in most cases continuous, numeric variable per time interval, representing the patient's most recently observed value in that interval, as well as the minimum, maximum, average. The patterns and codes used for data extraction were compiled in cooperation with SMEs.

\begin{minipage}{\textwidth}
  \begin{minipage}[b]{0.49\textwidth}
    \centering
    \includegraphics[width=1.1\textwidth]{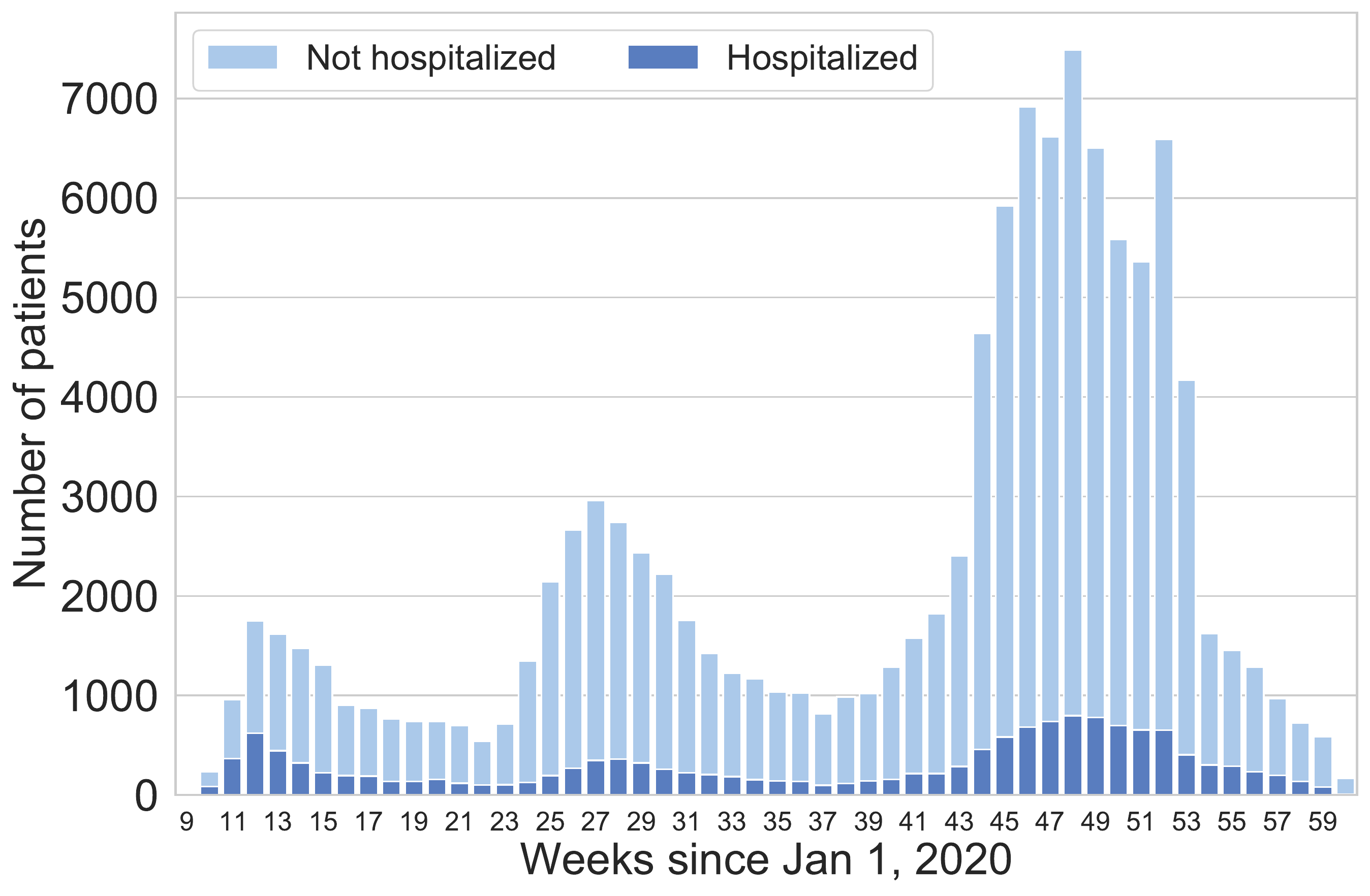}
    \captionof{figure}{Number of COVID-19 cases by number of weeks since January 1, 2020.}
    \label{fig:case_history}
  \end{minipage}
  \hfill
  \begin{minipage}[b]{0.49\textwidth}
    \centering
    \setlength\extrarowheight{-5.0pt}
    \resizebox{0.58\textwidth}{!}{
    \begin{tabular}{llllll}
    \hline
    & \\[-0.2cm]
     & \boldmath {\bf Number of patients} \\ \hline   
        & \\[-0.2cm]
        {\it Gender} &  \\ 
        \hspace{3mm}{\it Female} & 63101 (55.38\%) \\ 
        \hspace{3mm}{\it Male} & 50827 (44.61\%) \\
        \hspace{3mm}{\it Unknown} & 13 (0.01\%) \\
        \hline
        & \\[-0.2cm]
        {\it Age} &  \\ 
        \hspace{3mm}{\it Below 10} & 2612 (2.29\%) \\ 
        \hspace{3mm}{\it 10 to 20} & 7852 (6.89\%) \\ 
        \hspace{3mm}{\it 20 to 30} & 17206 (15.10\%) \\ 
        \hspace{3mm}{\it 30 to 40} & 17401 (15.27\%) \\ 
        \hspace{3mm}{\it 40 to 50} & 16744 (14.70\%) \\ 
        \hspace{3mm}{\it 50 to 60} & 19019 (16.69\%) \\ 
        \hspace{3mm}{\it 60 to 70} & 16417 (14.41\%) \\ 
        \hspace{3mm}{\it 70 to 80} & 10116 (8.88\%) \\ 
        \hspace{3mm}{\it Above 80} & 6222 (5.46\%) \\ 
        \hspace{3mm}{\it Unknown} & 352 (0.31\%) \\
        \hline
        & \\[-0.2cm]
        {\it State} &  \\ 
        \hspace{3mm}{\it Ohio} & 50948 (44.71\%) \\ 
        \hspace{3mm}{\it Louisiana} & 50638 (44.44\%) \\
        \hspace{3mm}{\it Illinois} & 5213 (4.58\%) \\
        \hspace{3mm}{\it Others (each $<$ 3.00\%)} & 7142 (6.27\%) \\
        \hline
        & \\[-0.2cm]
    {\bf COVID-19 cohort} & 113941 (100.00\%) \\
        \hline
    \end{tabular}
    }
      \captionof{table}{Basic statistics of our COVID-19 cohort in terms of patient gender, age, and location.}
      \label{table:stasboolean2}
    \end{minipage}
    \vspace{0.5cm}
  \end{minipage}
  
In order to determine if a patient had been hospitalized in a given time interval, we examined the patient interactions with the healthcare providers, as captured in the patient's \textit{encounter} data. A patient was considered hospitalized in a time interval \textit{T}, if they had an encounter with (i) encounter type {\it inpatient}, {\it hospital emergency room visit} or {\it hospital encounter}, (ii) admission date within \textit{T}, AND (iii) duration above 24 hours.
To exclude cases potentially hospitalized due to reasons other than COVID-19, patients with a hospitalization within the 4 weeks prior to the COVID-19 diagnoses were removed. After this cleanup step, patients with a hospitalization in the 4 weeks following the COVID-19 diagnosis were considered hospitalized due to COVID-19; we will refer to these patients simply as ``hospitalized''. Interestingly, 79\% of the hospitalized patients were admitted within 4 days following the COVID-19 diagnosis. The prediction target of our models was based on these COVID-19 hospitalization labels. It must be noted that some patients may be hospitalized in other institutions not part of the Explorys network, producing noise in form of false negatives. A possible reason could be bed unavailability due to sudden spikes in demand. The temporal distribution of hospitalized and non-hospitalized cases is illustrated in Figure~\ref{fig:case_history}.

\subsection*{Data preparation and feature extraction}

Possible bias in the data was removed by discarding all features related to ethnicity, race, medical insurance type, location and any other available information not related to the medical conditions of the patient. Used feature types were demographics (age, gender), clinical presentation (symptoms, vitals, other medical conditions, comorbidities) and lab measurements. These categorical and continuous variables were associated to specific time intervals as previously described. We removed patients in case age information was unavailable or if all other features were empty. Missing values were replaced using data imputation methods, and features were encoded (e.g., one hot encoding for the age groups),  after randomly splitting the data as 70\% train set and 30\% test set. More precisely, missing values for Boolean features were imputed with false, while continuous features were imputed with median substitutions learned on the train set. After feature imputation, we performed data normalization using average and standard deviation computed on the train set. For comorbidities and chronic conditions, mentions over all the available time intervals were aggregated using disjunction ("OR") into a single Boolean feature (condition is present). For the other feature types, in order to avoid data leakage\cite{KaufmanACM2012}, we selected only the time intervals from 14 days prior to the diagnosis up to the hospitalization ({\it period of interest}) with an upper limit
of 28 days after COVID-19 diagnosis. Labs and vitals were kept with the corresponding time information, while all other Boolean medical conditions including symptoms were aggregated over the {\it period of interest}. We are aware that from a medical perspective the knowledge of whether a symptom was present for one or more consecutive days would be important to build the model, unfortunately our data did not consistently contain this type of information. Finally, for some conditions (i.e., stroke) we used the time intervals to distinguish between conditions belonging to the medical history of the patient ("past") or to the {\it period of interest}.
After the preprocessing steps described above
the cohort
comprised a total number of patients [110996], non-hospitalized [97082, \%87] and hospitalized [13914, \%13].
Since most patients were hospitalized within 4 days from the COVID-19 diagnosis date and given the sparsity of the features over time, we explored two approaches that combine the two data modalities (time series and tabular) at different stages in the model.
In the first approach, the features from the two modalities
were concatenated ({\it early fusion}\cite{MultimodalDL}) in an ordered feature vector
$\mathcal{X} = [\mathcal{X}_1, \mathcal{X}_2]$
where $\mathcal{X} \in \mathcal{R}^{n\times k}$
with $n$ and $k$ the total numbers of patients
and features, respectively. Data modality $\mathcal{X}_1 \in \mathcal{R}^{n\times h}$
contains tabular data such as gender, age groups, comorbidities and other Boolean medical conditions, for a total of $h$ time-independent features. Data modality $\mathcal{X}_2 \in \mathcal{R}^{n\times m \times t}$ 
contains $m$ features collected over
$t$ time intervals, such as lab values and vitals. Concatenation of the modalities was performed after reshaping the second modality to a two dimensional matrix.
In the second approach, the modalities $\mathcal{X}_1$ and $\mathcal{X}_2 $ are fed into two different architectures, and their hidden representations are subsequently merged in the model ({\it model fusion}\cite{MultimodalDL}). Upon the described preprocessing, the total number of features $k = m \times t + h$ is $k = 1573$ with $h = 77$, $m = 88$ and $t = 17$.

\subsection*{Statistics}
\label{sec:stats}


    \begin{table}[h!]
    \centering
    \setlength\extrarowheight{-2.5pt}
    \resizebox{0.85\textwidth}{!}{
    \begin{tabular}{llllll}
    \hline
    {\bf Feature} & \boldmath $H_0$ (97082)& \boldmath $H_1$ (13914)& {\bf Feature} & \boldmath $H_0$ (97082)& \boldmath $H_1$ (13914)\\    \hline
        {\it Hypertension} & 32.22 & 72.78$^{**}$ & 
    {\it Sepsis} & 0.67 & 7.21$^{**}$ \\ 
        {\it Male} & 43.82 & 49.55$^{**}$ & 
    {\it Vomiting} & 1.88 & 6.41$^{**}$ \\ 
        {\it Pneumonia} & 5.03 & 48.83$^{**}$ & 
    {\it Age $\in [30, 40)$} & 16.87 & 6.08$^{**}$ \\
        {\it Diabetes} & 13.95 & 43.40$^{**}$ & 
    {\it Abdominal pain} & 2.41 & 6.06$^{**}$ \\ 
        {\it Hypoxia} & 1.76 & 38.42$^{**}$ & 
    {\it Gastroesophageal reflux} & 2.12 & 5.94$^{**}$ \\ 
        {\it Dyspnea} & 10.05 & 36.40$^{**}$ & 
    {\it Cerebrovascular disease (present)} & 0.66 & 4.76$^{**}$ \\ 
        {\it Heart disease} & 5.25 & 33.76$^{**}$ & 
    {\it Embolism} & 0.56 & 4.71$^{**}$ \\ 
        {\it Hypoxemia} & 1.31 & 31.36$^{**}$ & 
    {\it Dementia} & 0.54 & 4.64$^{**}$ \\ 
        {\it Nicotine dependence} & 14.63 & 31.04$^{**}$ & 
    {\it Headache} & 7.15 & 4.43$^{**}$ \\ 
        {\it Chronic kidney disease} & 5.29 & 27.17$^{**}$ & 
    {\it Age $\in [20, 30)$} & 17.05 & 3.62$^{**}$ \\ 
        {\it Cough} & 25.70 & 24.21$^{**}$ & 
    {\it Dizziness} & 1.32 & 3.31$^{**}$ \\ 
        {\it Coronary heart disease} & 6.15 & 23.90$^{**}$ & 
    {\it Acute respiratory distress syndrome} & 0.54 & 3.19$^{**}$ \\ 
        {\it Cancer} & 10.36 & 23.57$^{**}$ & 
    {\it Syncope} & 0.54 & 3.07$^{**}$ \\ 
        {\it Age $\in [60, 70)$} & 13.03 & 23.45$^{**}$ &
    {\it Cerebrovascular accident (present)} & 0.33 & 2.99$^{**}$ \\ 
        {\it Arrhythmia} & 3.44 & 22.78$^{**}$ & 
    {\it Pulmonary embolism} & 0.28 & 2.68$^{**}$ \\ 
        {\it Chronic obstructive lung disease} & 5.01 & 22.49$^{**}$ & 
    {\it Myalgia} & 3.43 & 2.64$^{**}$ \\ 
        {\it Age $\in [70, 80)$} & 6.85 & 20.60$^{**}$ & 
    {\it Fibromyalgia} & 1.69 & 2.62$^{**}$ \\ 
        {\it Renal failure syndrome} & 1.60 & 20.40$^{**}$ & 
    {\it Pregnancy} & 2.26 & 2.49 \\ 
        {\it Asthma} & 12.96 & 19.46$^{**}$ & 
    {\it Delirium} & 0.14 & 2.35$^{**}$ \\ 
        {\it Fever} & 12.32 & 18.89$^{**}$ & 
    {\it Seizure} & 0.42 & 2.34$^{**}$ \\ 
        {\it Fatigue} & 7.13 & 17.94$^{**}$ & 
    {\it Cirrhosis} & 0.62 & 2.27$^{**}$ \\ 
        {\it Cerebrovascular disease (past)} & 4.63 & 17.65$^{**}$ & 
    {\it Septic shock} & 0.17 & 1.32$^{**}$ \\ 
        {\it Age $\in [80, 90)$} & 3.25 & 17.55$^{**}$ &
    {\it Sore throat} & 3.97 & 1.04$^{**}$ \\ 
        {\it Age $\in [50, 60)$} & 16.75 & 17.50 &
    {\it Pulmonary hypertension} & 0.14 & 0.86$^{**}$ \\ 
        {\it Hyperlipidemia} & 5.16 & 17.13$^{**}$ & 
    {\it Age $\in [10, 20)$} & 7.94 & 0.83$^{**}$ \\
        {\it Acute renal failure syndrome} & 1.31 & 17.08$^{**}$ & 
    {\it Angina pectoris} & 0.17 & 0.72$^{**}$ \\ 
        {\it Chest pain} & 5.01 & 14.52$^{**}$ & 
    {\it HIV} & 0.38 & 0.66$^{**}$ \\ 
        {\it Tachycardia} & 1.92 & 13.22$^{**}$ & 
    {\it Rhinorrhea} & 1.35 & 0.52$^{**}$ \\ 
        {\it Acute disease of cardiovascular system} & 1.60 & 11.13$^{**}$ & 
    {\it Viral URI with cough} & 0.85 & 0.51$^{**}$ \\ 
        {\it Anemia} & 1.79 & 10.16$^{**}$ & 
    {\it Hemoptysis} & 0.11 & 0.49$^{**}$ \\ 
        {\it Age $\in [40, 50)$} & 15.61 & 10.09$^{**}$ &
    {\it Pneumothorax} & 0.09 & 0.36$^{**}$ \\ 
        {\it Heart failure} & 1.07 & 9.43$^{**}$ & 
    {\it Age $< 10$} & 2.65 & 0.27$^{**}$ \\ 
        {\it Atrial fibrillation} & 1.30 & 8.98$^{**}$ & 
    {\it Dry cough} & 0.25 & 0.24 \\ 
        {\it Nausea} & 4.14 & 8.81$^{**}$ & 
    {\it Thromboembolic disorder} & 0.05 & 0.22$^{**}$ \\ 
        {\it Diarrhea} & 3.54 & 8.78$^{**}$ & 
    {\it Productive cough} & 0.18 & 0.21 \\ 
        {\it Hypokalemia} & 1.06 & 8.56$^{**}$ & 
    {\it Myocarditis} & 0.02 & 0.16$^{**}$ \\ 
        {\it Cerebrovascular accident (past)} & 1.80 & 8.47$^{**}$ & 
    {\it Acute hepatic failure} & 0.01 & 0.07$^{**}$ \\ 
        {\it Chronic liver disease} & 3.83 & 7.96$^{**}$ & 
     & & \\ 
    \hline
    \end{tabular}
    }
    \caption{Demographic information, pre-existing comorbidities and acute conditions in \% on COVID-19 positive patients.
    Comparison of non-hospitalized ($H_0$) and hospitalized ($H_1$) groups is performed with a $\chi^2$ contingency analysis with post-hoc correction (Benjamini–Hochberg procedure) 
    is shown in the table with two asterisks when p-values $<0.001$. Gender is one for males and zero for females.}
    \label{table:stasboolean}
    \end{table}

Statistical analysis was performed to compare the hospitalized ($H_1$) and non-hospitalized ($H_0$) populations and compute the significance on the rejection of the null hypothesis that {\it there is no relationship
between predictor and outcome}. Demographic information, symptoms, comorbidities and other medical conditions are presented (as \%) in Table~\ref{table:stasboolean}, while median and IQR values for lab measurements and vitals are presented in Table~\ref{table:stastime}. Comparison of the $H_1$ and $H_0$ groups was performed with a $\chi^2$ contingency analysis for
the Boolean feature and Mann–Whitney U test for the numerical ones. In our analysis we used post-hoc correction with the Benjamini–Hochberg procedure. Two asterisks were assigned when p-values are $<0.001$. 

When comparing the percentages of symptoms and comorbidities reported in Table~\ref{table:stasboolean} to existing statistics, it should be noted that these numbers were calculated on the preprocessed dataset upon removal of all features collected after hospitalization. Large variability is generally observed across publications, likely due to differences in the cohorts, the way data were collected and the time the patient met the caregiver in their disease progression journey (e.g. at triage vs hospitalization). In the laboratory-confirmed US cohort described in \cite{usa1mcases}, cough is reported in about 50\% and fever in 43\% of the 28.3\% of the patients for which symptoms are actually known. Several of the conditions more commonly found in $H_1$ in Table~\ref{table:stasboolean} are in agreement with literature findings, either as conditions correlated to severe disease progression (e.g. pneumonia, hypoxia, hypoxemia) or comorbidities and risk factors (e.g. hypertension, diabetes, nicotine dependence, chronic kidney disease).
Our analysis shows increasing hospitalization 
admissions associated with older age: eighty year old patients are 4 times higher in $H_1$ than $H_0$, and 2 times higher for
seventy and sixty year old patients. 

Regarding the lab values reported in Table~\ref{table:stastime}, it should be noted that these values are recorded before hospital admission for $H_1$ patients (so relatively early in the patient journey). Also, as previously mentioned, since data are sourced from various institutions, high variability amongst measurements and reference values can contribute to noise. Median values for Albumin (ALB), Platelet count (PLT) and SpO2 are slightly lower in $H_1$. Median values of Aspartate aminotransferase (AST), Serum ferritin (Ferritin), Lactate dehydrogenase (LDH), high-sensitive C-reactive protein (CRP), White blood cell count (WBC) are higher in $H_1$ (similar findings of \cite{Lancet1}).


    \begin{table}
    \centering
    \setlength\extrarowheight{-1.5pt}
    \resizebox{\textwidth}{!}{
    \begin{tabular}{llllllllll}
    \hline
    {\bf Feature} & {\bf Count} & \boldmath $H_0$ (97082)& {\bf Count} & \boldmath $H_1$ (13914)& {\bf Feature} & {\bf Count} & \boldmath $H_0$ (97082)& {\bf Count} & \boldmath $H_1$ (13914)\\    \hline
        {\it HR (beats/min)} & 49366& 85.00 [75.00-96.00] & 14469& 84.00 [73.00-95.00]$^{**}$ & 
    {\it Glucose (mg/dL)} & 7815& 114.00 [97.00-156.00] & 5286& 122.00 [103.25-167.00]$^{**}$ \\ 
        {\it SBP (mmHg)} & 46500& 130.00 [118.00-141.00] & 14070& 129.00 [115.00-143.00]$^{**}$ & 
    {\it Neutrophils (10$^3$/$\mu$L)} & 4163& 3.62 [2.55-5.28] & 5039& 4.34 [3.02-6.38]$^{**}$ \\ 
        {\it DBP (mmHg)} & 46494& 78.00 [69.00-85.00] & 14064& 71.00 [63.00-81.00]$^{**}$ & 
    {\it CRP (mg/dL)} & 3075& 5.30 [1.56-12.60] & 4894& 7.10 [3.10-13.30]$^{**}$ \\ 
        {\it T ($^\circ$C)} & 50027& 36.90 [36.70-37.20] & 13696& 37.00 [36.70-37.40]$^{**}$ & 
    {\it Ferritin (ng/mL)} & 2528& 386.00 [160.00-917.00] & 4143& 475.00 [213.00-964.50]$^{**}$ \\ 
        {\it BMI (Kg/m$^2$)} & 40873& 29.12 [24.70-34.45] & 12993& 30.52 [25.93-36.50]$^{**}$ & 
    {\it LDH (U/L)} & 2227& 318.00 [223.00-487.50] & 3888& 335.00 [244.00-477.25]$^{**}$ \\ 
        {\it RR (breaths/min)} & 39226& 18.00 [16.00-19.00] & 12809& 18.00 [18.00-20.00]$^{**}$ & 
    {\it CK (U/L)} & 2236& 113.00 [61.00-271.00] & 3362& 132.00 [66.00-288.00] \\ 
        {\it Hb (g/dL)} & 12417& 13.20 [11.90-14.40] & 11053& 12.90 [11.50-14.20]$^{**}$ & 
    {\it D-dimer (ng/mL)} & 2120& 726.50 [410.00-1539.25] & 3296& 870.00 [520.00-1678.50]$^{**}$ \\ 
        {\it Crea (mg/dL)} & 12369& 0.95 [0.80-1.23] & 10898& 1.05 [0.80-1.46]$^{**}$ & 
    {\it PCT (ng/mL)} & 2042& 0.15 [0.06-0.45] & 3292& 0.14 [0.08-0.34] \\ 
        {\it PLT (10$^3$/$\mu$L)} & 11658& 210.00 [165.00-262.00] & 10662& 200.00 [157.00-256.00]$^{**}$ & 
    {\it hsTnT (ng/mL)} & 2019& 0.03 [0.01-0.07] & 2452& 0.03 [0.01-0.06] \\ 
        {\it WBC (10$^3$/$\mu$L)} & 11566& 5.83 [4.50-7.81] & 10661& 6.31 [4.74-8.54]$^{**}$ & 
    {\it aPTT (s)} & 1161& 30.30 [27.40-34.80] & 1930& 29.80 [26.90-33.60]$^{**}$ \\ 
        {\it K (mmol/L)} & 11831& 4.00 [3.70-4.30] & 10481& 3.90 [3.60-4.30] & 
    {\it Eosinophils (10$^3$/$\mu$L)} & 2012& 0.08 [0.04-0.14] & 1700& 0.08 [0.04-0.14] \\ 
        {\it BUN (mg/dL)} & 11682& 15.00 [11.00-21.00] & 10327& 17.00 [12.00-27.00]$^{**}$ & 
    {\it Basophils (10$^3$/$\mu$L)} & 1605& 0.03 [0.03-0.05] & 1652& 0.03 [0.03-0.05] \\ 
        {\it ALB (g/dL)} & 10794& 3.80 [3.30-4.20] & 10075& 3.60 [3.20-4.00]$^{**}$ & 
    {\it PH} & 413& 7.38 [7.31-7.44] & 673& 7.42 [7.38-7.46]$^{**}$ \\ 
        {\it AST (U/L)} & 10416& 30.00 [21.00-45.00] & 9683& 34.00 [24.00-50.00]$^{**}$ & 
    {\it CO2 (mmHg)} & 422& 38.00 [32.00-46.00] & 655& 36.10 [31.40-42.00] \\ 
        {\it Monocytes (10$^3$/$\mu$L)} & 10149& 0.50 [0.40-0.70] & 9539& 0.50 [0.35-0.70] & 
    {\it HCO3 (mmHg)} & 406& 22.10 [19.00-25.30] & 646& 23.50 [21.00-26.00]$^{**}$ \\ 
        {\it Bilirubin (mg/dL)} & 10023& 0.50 [0.30-0.70] & 9519& 0.50 [0.40-0.70]$^{**}$ & 
    {\it O2 (mmHg)} & 399& 83.70 [64.00-114.50] & 598& 79.40 [62.82-107.00] \\ 
        {\it Lymphocytes (\%)} & 10201& 20.90 [13.10-30.10] & 9339& 16.30 [10.70-23.50]$^{**}$ & 
    {\it WBC (U/hpf)} & 308& 6.00 [6.00-11.00] & 509& 11.00 [6.00-26.40]$^{**}$ \\ 
        {\it ALT (U/L)} & 9935& 25.00 [17.00-38.00] & 9306& 25.00 [16.00-39.00] & 
    {\it PTT (s)} & 375& 13.70 [12.80-15.45] & 391& 13.80 [13.10-15.40] \\ 
        {\it Lymphocytes (10$^3$/$\mu$L)} & 9769& 1.20 [0.80-1.70] & 9133& 1.00 [0.70-1.40]$^{**}$ & 
    {\it Bilirubin} & 199& 1.00 [1.00-2.00] & 173& 1.00 [1.00-1.00] \\ 
        {\it SpO2 (\%)} & 13798& 98.00 [96.00-99.00] & 7131& 96.00 [94.00-98.00]$^{**}$ & 
    {\it Lactate (mmol/L)} & 50& 1.10 [0.72-1.60] & 60& 1.00 [0.80-1.32] \\ 
        {\it Na (mmol/L)} & 5944& 138.00 [135.00-140.00] & 6414& 137.00 [134.00-139.00]$^{**}$ & 
    {\it mean BD (mmHg)} & 69& 90.00 [79.00-102.00] & 39& 89.00 [80.00-97.50] \\ 
        {\it HR pulse oximetry (\%)} & 29854& 98.00 [97.00-99.00] & 5742& 96.00 [94.00-98.00]$^{**}$ & 
     & & & & \\ 
    \hline
    \end{tabular}
    }
    \caption{Median and IQR of the time dependent features reported in the EHR close to the COVID-19 diagnosis. Comparison of non-hospitalized ($H_0$) and hospitalized ($H_1$) groups is performed with Mann–Whitney U test with post-hoc correction (Benjamini–Hochberg procedure) is shown in the table
    with two asterisks when p-values $<0.001$.}
    \label{table:stastime}
    \end{table}

\subsection*{Training and explainability}
\label{section:model}

Prediction of hospitalization up to 28 days after
COVID-19 diagnosis was modelled as a binary 
classification ($H_1$ hospitalized, $H_0$ not hospitalized). For {\it early fusion} approach we 
trained the following models:
Random Forest classifier (RF) \cite{rf1},
Extra Trees classifier (ET) \cite{rf1},
fully connected neural network (NN) \cite{deeplearningreview}
and Auto AI architectures (Auto Keras \cite{autokeras}).
For the {\it model fusion} approach we trained
an architecture combining NN for the time-independent modality
and temporal convolutional neural network (T-CNN) \cite{deeplearningreview} for the time-dependent one.
The two models were combined together with additional NN layers. The T-CNN is a convolutional neural network consisting of 1-D temporal convolutions.
During training we performed 3-fold cross validation
and selected the best performing (best $F_1$-score) model
on the validation set. Splitting
of the dataset into train, validation, and test
sets was performed several times varying the seed
for the splitting. Precision, recall
and $F_1$-score were collected by computing
average and standard deviation over
all iterations. RF and ET models
were trained with a randomized 
parameter optimization to search
for the optimal hyperparameters 
such as number of trees, number of features 
to consider for the best split, maximum tree depth, minimum number of samples required to split an internal node, the minimum number of samples required to be at a leaf node either using the whole sample
or bootstrapping.
For the AutoKeras models we have varied the number
of iterations and the number of epochs
and let the library to find the best
performing model.
NN and T-CNN were optimized using mini batch Stochastic Gradient Descend with momentum, learning rate was varied with a step scheduler
and dropout used to reduce overfitting.
Comparing to the architecture 
trained by the Auto AI model with 
three dense layers followed by $ReLu$
activation functions our implementations of
NN (T-CNN) contained more (less) parameters.
In all models we have defined a custom loss
and weighted random sampling of the training data
to take into account the high class unbalance.

First, we built the models using all available features to have baselines ({\it all features} use case). Then, to test model adaptability to different use cases, we removed selected lists of features.  To emulate general practitioner settings (GP), where diagnoses of acute conditions are not available and the results of some very specific tests not accessible, we removed the following: {\it pneumonia, hypoxia, hypoxemia,
myocarditis, all sepsis related conditions,
acute respiratory distress syndrome,
heart disease, pneumothorax,
acute renal failure syndrome, thromboembolic disorder,
acute hepatic failure, pulmonary embolism,
renal failure syndrome, embolism,
acute disease of cardiovascular system and lab tests such as D-dimer and hsTnT.} This {\it GP} use case was designed under SMEs guidance. Finally, to test the model predictive power over time, we removed all
available features in the day prior to the
hospitalization for hospitalized patients ({\it one day before} use case). This drastically reduces the availability of lab values and vitals. Average precision, recall, $F_1$-score are given in Table~\ref{tableres} for the {\it all feature} use case, the {\it GP} use case and the {\it one day before} use case, as first, second and third values in the series, respectively. NN and T-CNN architectures were frozen across all use cases.
In all described use cases, all analysed models achieved relatively high classification performance despite the unbalanced dataset. There is a minimal difference amongst the time dependent T-CNN model performance and the NN and the Auto Keras models for {\it all features} and the {\it GP} use cases. Drop of $F_1$-score when using a smaller number of features was limited to few percentages for all the models, at the exception of the T-CNN in {\it one day before} use case.  The observed similar performances of the models may be due to the presence of several predictive features in the dataset, most likely, more than the dropped ones.

\begin{table}[h!]
\centering
\resizebox{0.95\textwidth}{!}{
\vspace{0.5cm}
\begin{tabular}{lllllll}
\hline
\bf Model & \bf \boldmath P($H_0$) & \bf \boldmath R($H_0$) & \bf \boldmath F($H_0$) & \bf \boldmath P($H_1$) & \bf \boldmath R($H_1$) & \bf \boldmath F($H_1$)\\
\hline
\it RF & 0.97/0.97/0.96 & 0.98/0.97/0.98 & 0.98/0.97/0.97 & 0.83/0.79/0.84 & 0.82/0.81/0.74 & 0.83/0.80/0.79 \\
\it ET & 0.97/0.96/0.96 & 0.98/0.97/0.98 & 0.97/0.96/0.97 & 0.81/0.77/0.83 & 0.80/0.74/0.75 & 0.80/0.75/0.79 \\
\it NN & 0.98/0.97/0.97 & 0.96/0.95/0.95 & 0.97/0.96/0.96 & 0.74/0.72/0.72 & 0.85/0.82/0.79 & 0.79/0.77/0.75 \\
\it T-CNN & 0.98/0.98/0.97 & 0.96/0.95/0.94 & 0.97/0.96/0.96 & 0.75/0.71/0.67 & 0.83/0.84/0.83 & 0.79/0.77/0.74 \\
\it Auto Keras & 0.96/0.96/0.96 & 0.98/0.97/0.97 & 0.97/0.97/0.97 & 0.85/0.80/0.81 & 0.75/0.75/0.72 & 0.80/0.77/0.76 \\
\hline
\end{tabular}
}
\caption{Hospitalization prediction in terms of average
precision (P), recall (R), $F_1$-score (F) within 28 days after the COVID-19 diagnosis by using (i) all available features prior
to the hospitalization, (ii) carefully removing a selected list of features with guidance of SMEs,
and (iii) removing all available features up to
one day prior to the hospitalization.
}
\label{tableres}
\vspace{0.5cm}
\end{table}

\begin{figure}[h!]
\centering
\includegraphics[scale=0.23]{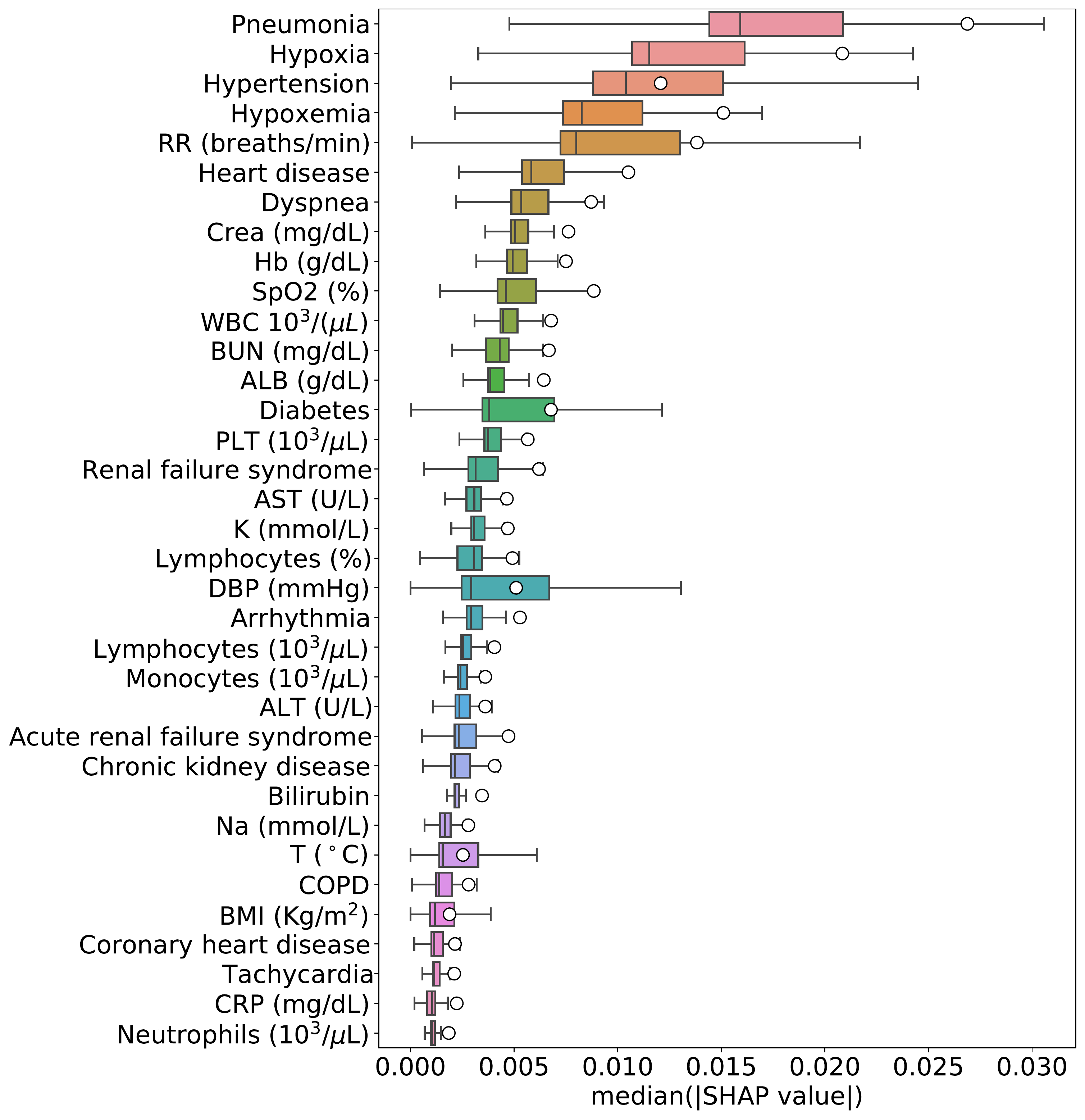}
\hspace{0.2cm}
\includegraphics[scale=0.23]{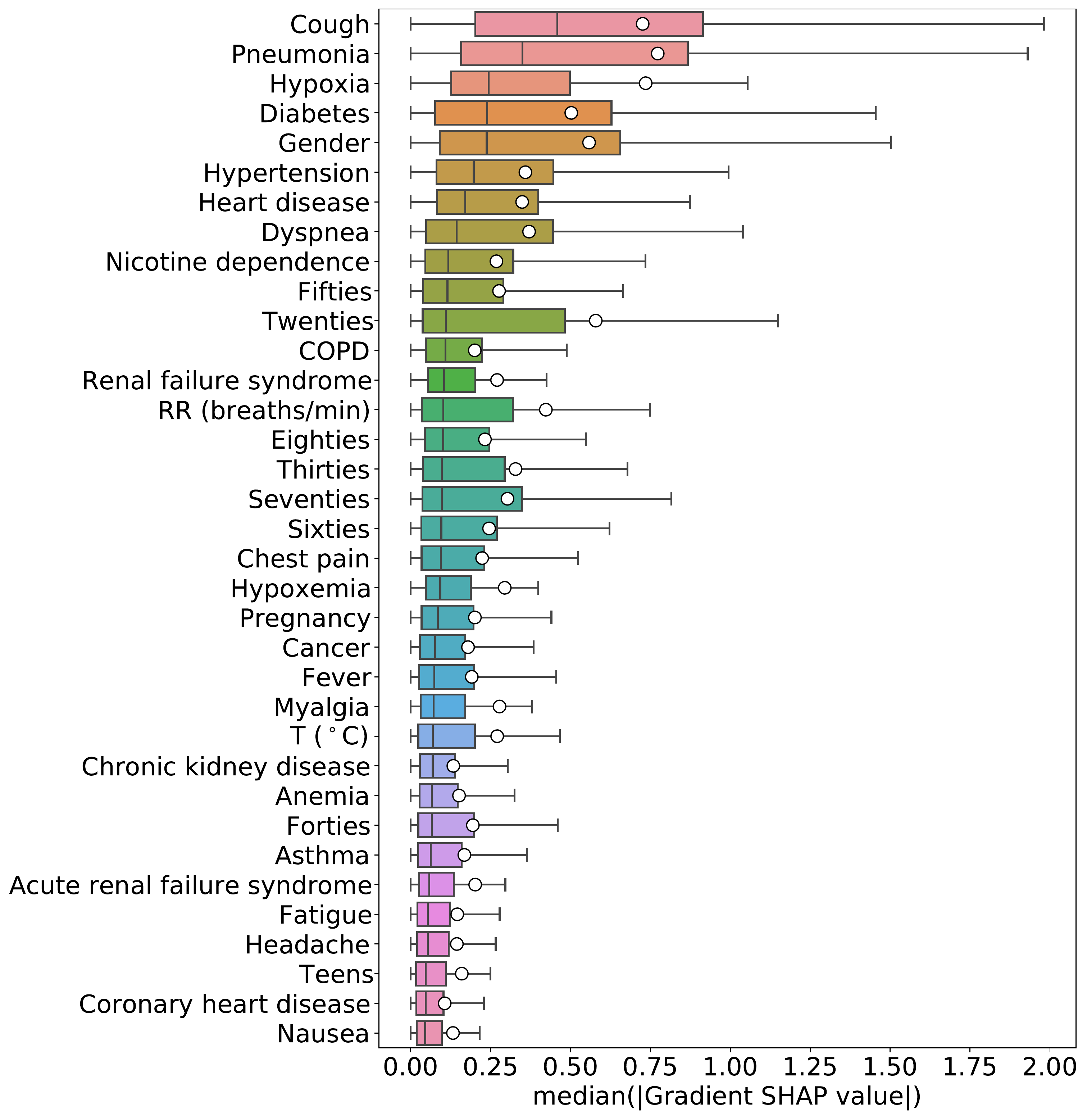}
\caption{Median SHAP values for the RF (left panel) and NN (right panel) computed on the test sets (across all splits) and using all available features. Corresponding interquartile ranges (IQR) are reported with colored boxes, $1.5 \times$ IQR ranges with whiskers and mean values with white circles.
}
\label{Fig:shap1}
\vspace{0.2cm}
\end{figure}

\begin{figure}[h!]
\centering
\hspace{0.6cm}
\includegraphics[scale=0.23]{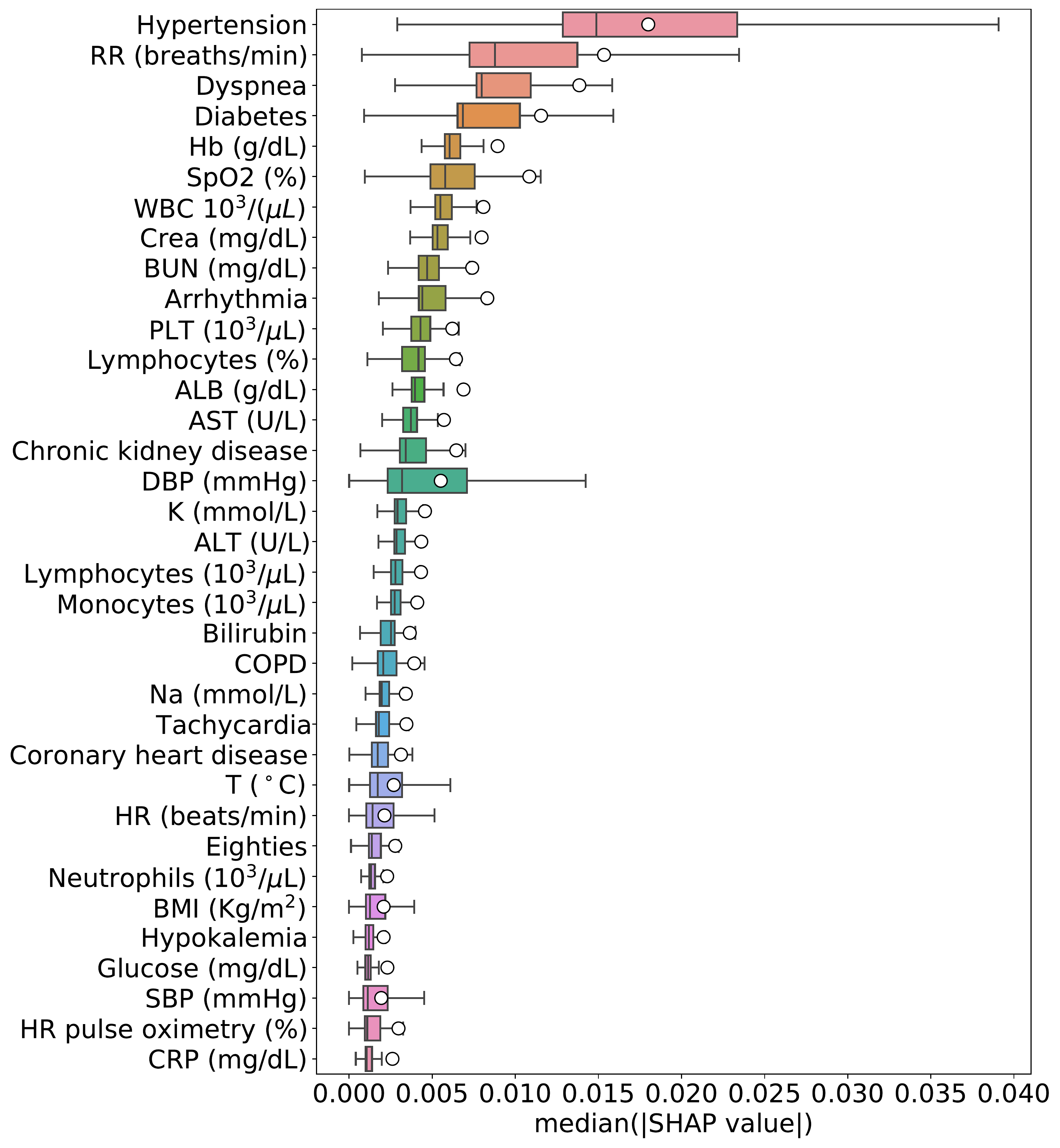}
\includegraphics[scale=0.23]{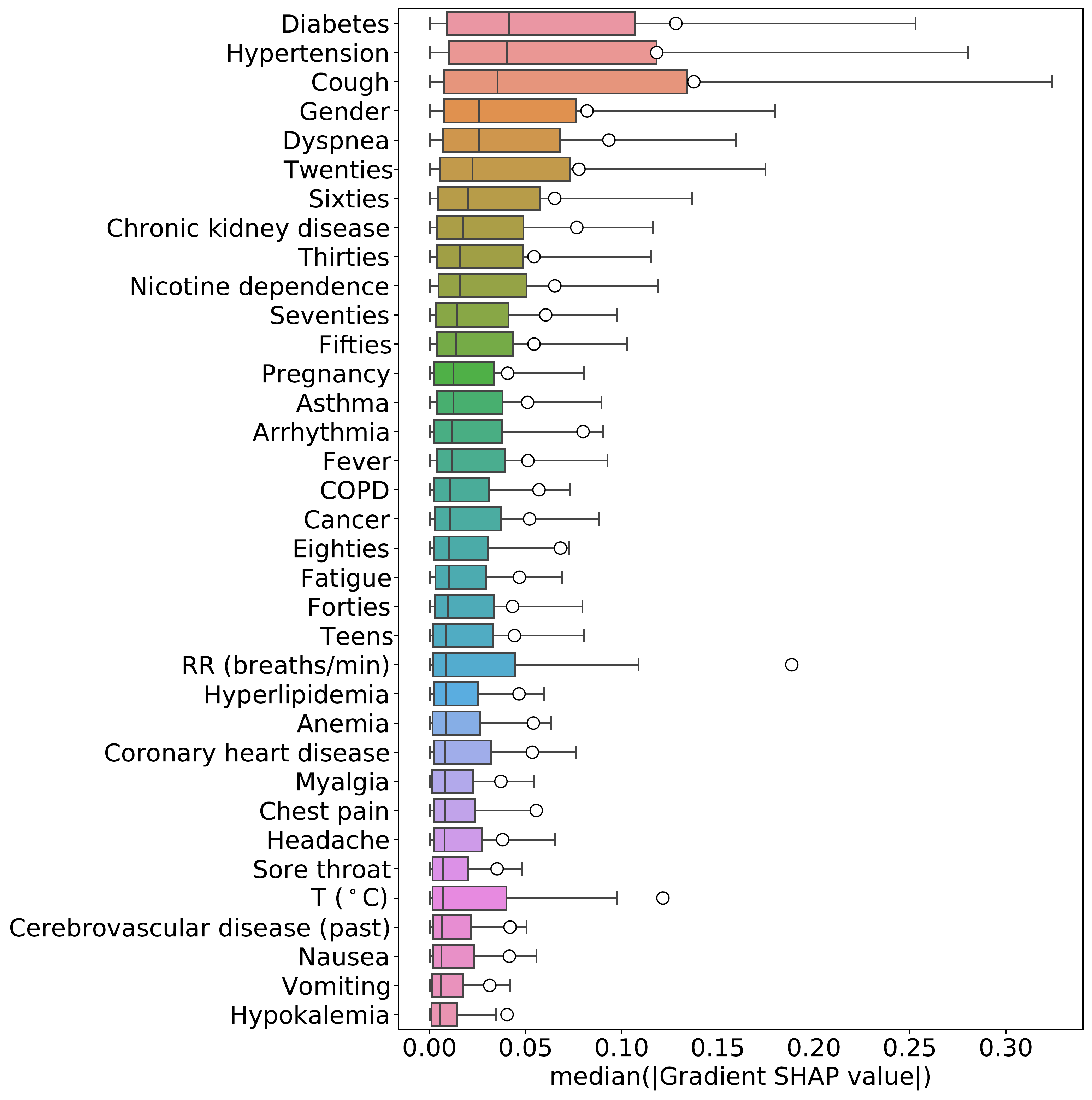}
\caption{Same as Figure~\ref{Fig:shap1}, but using a restricted number of available features. See text for further details.
}
\label{Fig:shap-gp}
\vspace{0.2cm}
\end{figure}

To determine which features have the largest predictive power and assess model interpretability, we explored feature importance methods. Although several studies \cite{Jimenez-SolemScienRep2021} used MDI importance\cite{GiniMDI} on datasets containing numerical features, we disregarded it, given its sensitivity to high-cardinality
features. In addition, commonly used libraries compute MDI-based importance values on the training set and therefore the extracted {\it important features} are not useful to make predictions that generalize on the test set. Instead, we focused on the SHAP method \cite{shap1} and report feature importance values computed on the test set. Regarding the NN and T-CNN models, SHAP values computed with the gradient based approximation are very similar to the ones computed using different not-gradient-based approximations. Therefore we did not regularize the gradients during training as suggested by \cite{gradientergul}. The top 35 SHAP values computed on the RF and NN trained with all available features are given in Figure~\ref{Fig:shap1} (corresponding plots for ET and CNN
have been omitted due to their similarity to the RF and the NN plots, respectively).
Median of the absolute SHAP values and corresponding interquartile ranges (IQR) are reported with colored boxes, $1.5 \times$ IQR ranges with whiskers and mean of the absolute values with white circles. Although some features overlap (12/35, 34\%), lab values and vitals are predominant in RF (60\%) while a minority in the NN model (6\%). This is even more accentuated after dropping the list of Boolean medical conditions in the GP use case (lab values 71\% in RF vs 6\% in NN, Figure~\ref{Fig:shap-gp}). When removing all features up to one day prior to the hospitalization, the number of lab values available to the model drops consistently and in this case the overlap between important features for the two models increases up to 25/35 (71\%).


\begin{figure}[h!]
\centering
\hspace{-0.68cm}
\includegraphics[scale=0.23]{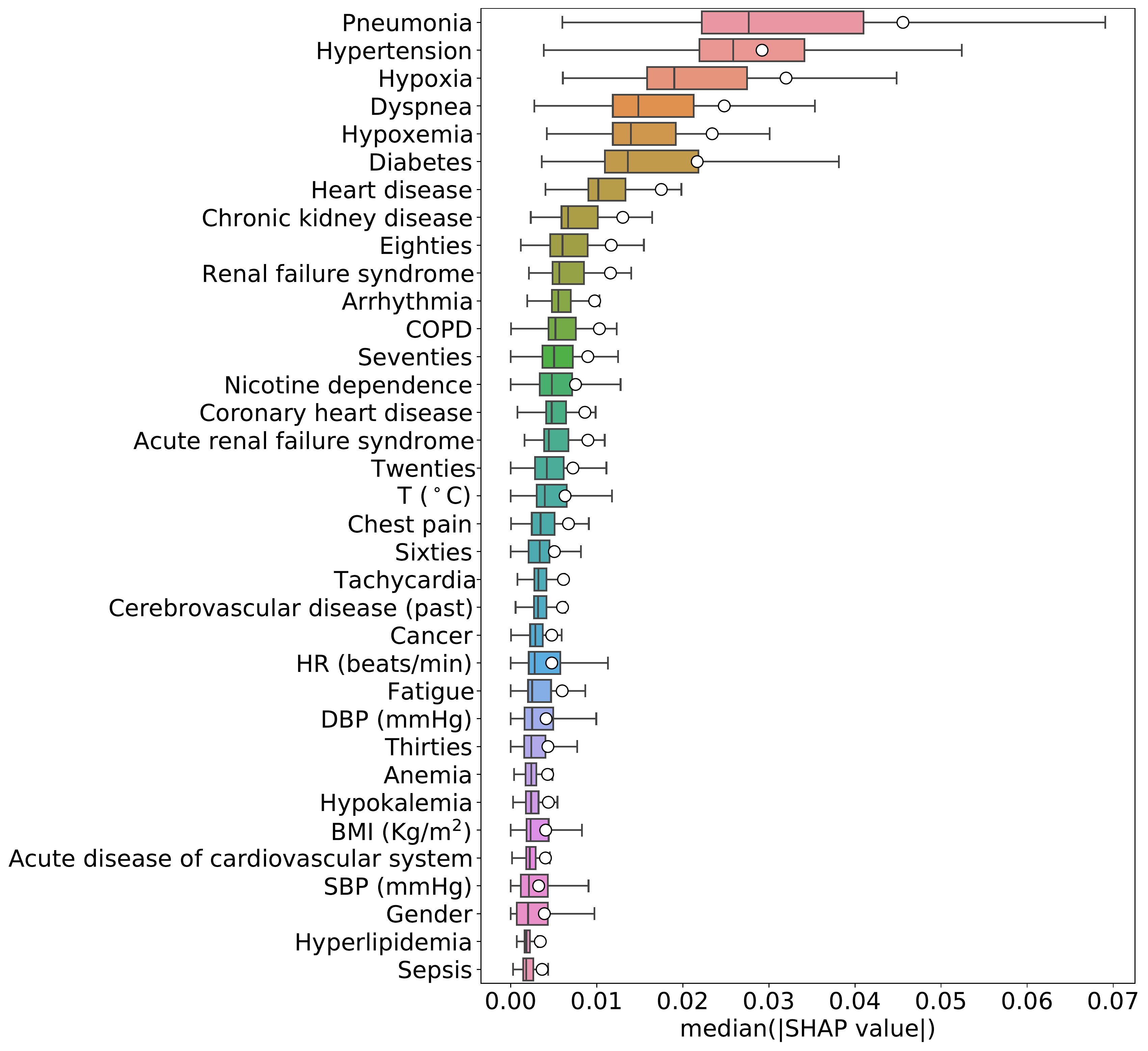}
\includegraphics[scale=0.23]{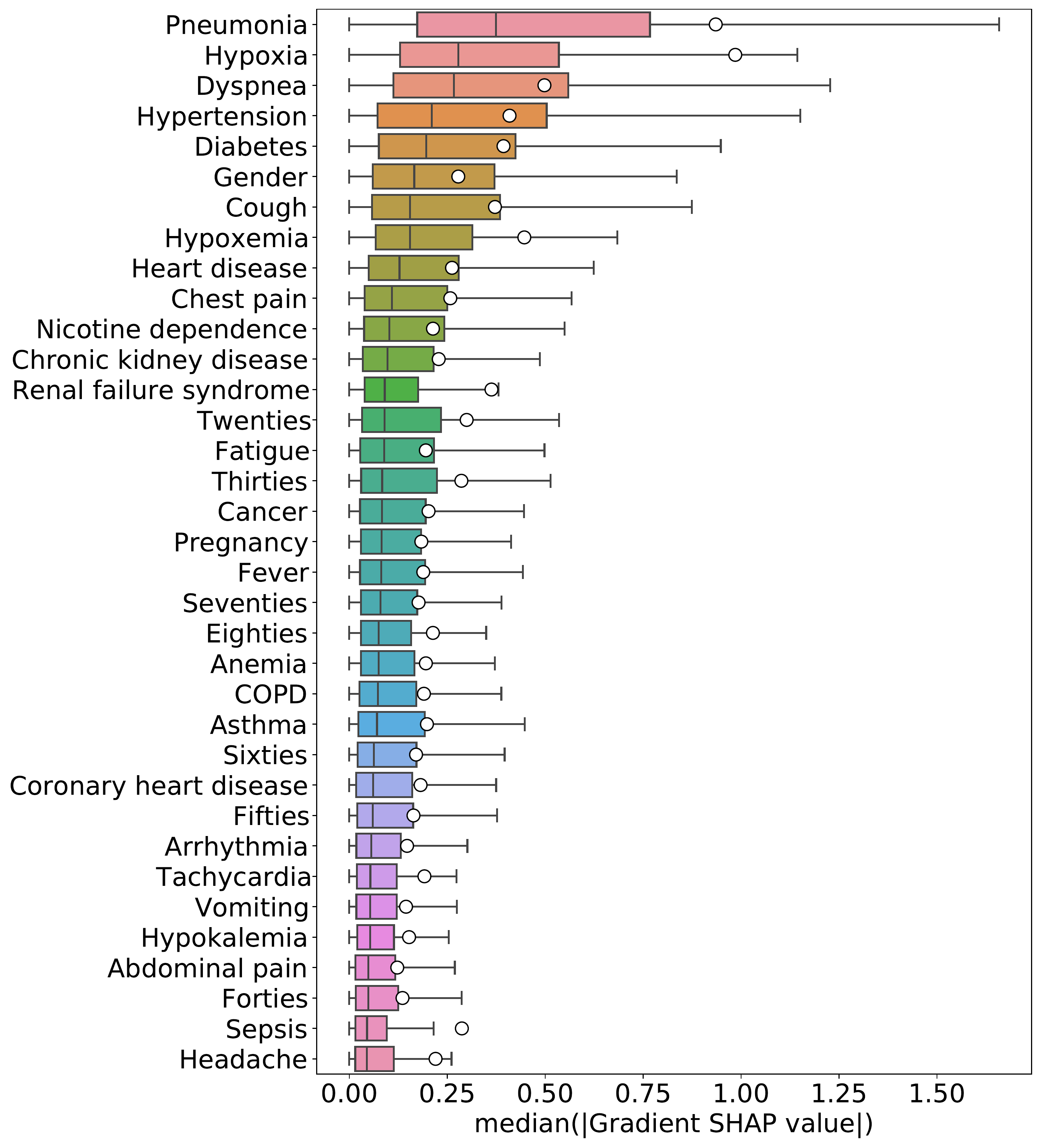}
\caption{Same as Figure~\ref{Fig:shap1}, but for RF and NN models after removing features.}
\label{Fig:shap-remove-one-day}
\vspace{0.2cm}
\end{figure}

For the Boolean conditions, a high level agreement is observed between the information of Table~\ref{table:stasboolean} and the feature importance results shown in Figures~\ref{Fig:shap1}, \ref{Fig:shap-gp}, \ref{Fig:shap-remove-one-day}. However, the statistical analysis is performed feature by
feature and is fundamentally different from the feature importance model described. The latter is based on a predictive model which takes into account
feature combination during the decision process
and hence does not require features to be
independent. In both cases, presence of relationship
from the statistical analysis or importance from
the SHAP method does not imply causality.
Reason for hospitalization comes from 
combination of several factors.

\section*{Conclusions}

We developed several AI models (RF, ET, NN, T-CNN) to predict hospitalization on a large cohort of COVID-19 patients (more than 110$k$) up to 28 days after the diagnosis, with competitive classification results on the test set. Despite the large data unbalance, using available features prior to the hospitalization, the models reach average 0.96-0.98 (0.75-0.85) precision, 0.96-0.98 (0.74-0.85) recall, and 0.97-0.98 (0.79-0.83) $F_1$-score across all split iterations for $H_0$ ($H_1$). Similar results were obtained in more restrictive scenarios with lower number of available features, even when features ranked as top by importance models were removed from the baseline model (GP use case).
We then systematically explored the SHAP feature importance method for all analysed models in all the use cases and observed an overall agreement with the results of the statistical analysis on the Boolean medical conditions.
However, the observed large variability of the SHAP importance score across models calls for a careful usage of the importance results, their evaluation
across different models and methods. When deriving feature importance, methods that can handle feature correlation like SHAP should be used
or analysis of feature correlation
carefully conducted.
Undesired dependencies as the ones
shown in this study need to be carefully evaluated
before adoption in high-stakes decisions \cite{ExplRudin2019,ExplFragile2019,Caruana2015}.

{\bf Limitations}

In this retrospective study we tried to adhere as much as possible to the PROBAST framework, in the constrained setting of this use case: hospitalization prediction requires a large dataset of hospitalized and non-hospitalized patients and therefore data sources of different provenience are needed (eg., GP office visits, emergency, hospital encounters). This means that if a patient left or suddenly came into the Explorys network for whatever reasons  (e.g., hospitalized in other hospitals or regions), outcomes and data may be missing. This may lead also to large variability in laboratory values and measurements. Our models were not tested on external datasets or in clinical settings due time constraints.

\makeatletter
\renewcommand{\@biblabel}[1]{\hfill #1.}
\makeatother

\bibliographystyle{unsrt}

\begin{thebibliography}{1}
\setlength\itemsep{-0.1em}


\bibitem{LancetV3J21} 
Artificial intelligence for COVID-19: saviour or saboteur?
The Lancet Digital Health 2012;3(1):e1-e66.

\bibitem{AroraFutVir2020} 
Arora N, Banerjee K, Narasu M L. The role of artificial intelligence in tackling COVID-19. Future Virology 2020;15(11):717-724.

\bibitem{WynantsBMJ2020} 
Wynants L, Van Calster B, Collins GS, Riley RD, Heinze G, Schuit E. Prediction models for diagnosis and prognosis of COVID-19: systematic review and critical appraisal. BMJ 2020;369.

\bibitem{MythGeneralisability} 
Futoma J, Simons M, Panch T, Doshi-Velez F, and Celi LA. The myth of generalisability in clinical research and machine learning in health care. The Lancet Digital Health, 2020;2(9):e489–e492.

\bibitem{PROBAST} 
Wolff RF, Moons KGM, Riley RD, et al. PROBAST: a tool to assess the risk of bias and applicability of prediction model studies. Ann Intern Med.2019;170:51-58.

\bibitem{TRIPOD} 
Collins GS, Reitsma JB, Altman DG, et al. Transparent Reporting of a multivariable prediction model for individual prognosis or diagnosis (TRIPOD): the TRIPOD statement. Ann Intern Med.2015;162:55-63.


\bibitem{ElifeScience} 
Hao B, Sotudian S, Wang T et al. Early prediction of level-of-care requirements in patients with COVID-19. eLife 2020;9:e60519.

\bibitem{Jimenez-SolemScienRep2021} 
Jimenez-Solem E, Petersen TS, Hansen C et al. Developing and validating COVID-19 adverse outcome risk prediction models from a bi-national European cohort of 5594 patients. Sci Rep 2021;11:3246.

\bibitem{JehiPlos} 
Jehi L, Ji X, Milinovich A, Erzurum S. et al. Development and validation of a model for individualized prediction of hospitalization risk in 4,536 patients with COVID-19. PloS one 2020;15(8):e0237419.

\bibitem{RichardsonJAMA2020} 
Richardson S, Hirsch JS, Narasimhan M, et al. Presenting characteristics, comorbidities, and outcomes among 5700 patients hospitalized with COVID-19 in the New York City area. JAMA. 2020;323(20):2052–2059.

\bibitem{Explorys} 
\url{https://www.ibm.com/products/explorys-ehr-data-analysis-tools}.

\bibitem{Ravizza} 
Ravizza S, Huschto T, Adamov A et al. Predicting the early risk of chronic kidney disease in patients with diabetes using real-world data. Nat Med 25, 57–59 (2019). 


\bibitem{KaufmanACM2012} 
Kaufman S, Rosset S, Perlich C, and Stitelman O. Leakage in data mining: formulation, detection, and avoidance. ACM Trans. Knowl. Discov. Data 6, 4, Article 15 (December 2012), 21 pages.

\bibitem{MultimodalDL} 
Ngiam J, Khosla A, Kim M, Nam J, Lee H, and Ng A. Multimodal deep learning. In Proceedings of International Conference on Machine Learning (ICML), 2011.


\bibitem{usa1mcases} 
Stokes EK, Zambrano LD, Anderson KN, et al. Coronavirus disease 2019 case surveillance — United States, January 22–May 30, 2020. MMWR Morb Mortal Wkly Rep 2020;69:759–765.


\bibitem{Lancet1} 
Zhou F, Yu T, Du R, et al. Clinical course and risk factors for mortality of adult inpatients with COVID-19 in Wuhan, China: a retrospective cohort study. {\it Lancet 2020 March 11}.

\bibitem{rf1} 
Breiman L. (2001). Random forests. {\it Machine learning}, 45(1):5–32.

\bibitem{deeplearningreview} 
LeCun, Y., Bengio, Y. \& Hinton, G. Deep learning. Nature 2015;521:436-444.

\bibitem{autokeras} 
Haifeng J, Qingquan S, and Xia H.
Auto-Keras: an efficient neural architecture search system.
Proceedings of the 25th ACM SIGKDD International Conference on Knowledge Discovery \& Data Mining,2019;1946-1956.

\bibitem{GiniMDI} 
Louppe G., Wehenkel L, Sutera A, and Geurts, P. Understanding variable importances in forests of randomized trees. In Advances in Neural Information Processing Systems 2013;431:439.

\bibitem{shap1} 
Lundberg SM and Lee S. A unified approach to interpreting model predictions. In {\it Advances in Neural Information Processing Systems} 2017;4768–4777.


\bibitem{gradientergul} 
Ross AS, Hughes MC, and Doshi-Velez F. Right for the right reasons: training differentiable models by constraining their explanations.
Proceedings of the Twenty-Sixth International Joint Conference on Artificial Intelligence (IJCAI-17), 2662--2670.


\bibitem{ExplRudin2019}
C. Rudin. Stop explaining black box machine learning models for high stakes decisions and use interpretable models instead. Nature Mach. Intell., 2019;1(5):206–-215.

\bibitem{ExplFragile2019} 
Ghorbani A, Abid A, Zou J. Interpretation of neural networks is fragile. AAAI. 2019; 17;33(01):3681--8.


\bibitem{Caruana2015}
Caruana R, Lou  Y, Gehrke J, Koch P,Sturm M, Elhadad N. Intelligible models for healthcare: predicting pneumonia risk and hospital 30-day readmission. In Proceedings of the 21st Annual SIGKDD International Conference on Knowledge Discovery and Data Mining, 2015;1721-1730.



\end{thebibliography}

\end{document}